\documentclass[preprint,12pt]{elsarticle}
\usepackage{amsmath,amssymb,amsfonts}
\usepackage{algorithmic}
\usepackage{graphicx}
\usepackage{textcomp}
\usepackage{tabularx}
\usepackage{booktabs}
\usepackage{multirow}
\usepackage{xcolor}
 
\usepackage{placeins}
\usepackage{csquotes}

\usepackage{amssymb}
\usepackage{amsmath}

\usepackage{url}
\journal{Neurocomputing}

\makeatletter
\def\ps@pprintTitle{%
  \let\@oddhead\@empty
  \let\@evenhead\@empty
  \let\@oddfoot\@empty
  \let\@evenfoot\@empty
}
\makeatother
 
\begin{document}

\title{Segment Anything Model Based
Context Guided Mamba Unet for Medical Image
Segmentation}

\begin{frontmatter}

\author[1]{Chenchen Wang}
\ead{wchenchen0808@gmail.com}

\address[1]{School of Computer Science and Informatics,
Cardiff University, Cardiff, UK}

\begin{abstract}
Medical image segmentation is crucial for clinical diagnosis and treatment planning, requiring both global contextual understanding and fine-grained structural precision. While the Segment Anything Model (SAM) demonstrates strong generalization capabilities, its direct application to medical images often lacks domain-specific adaptability. Meanwhile, State Space Models (SSMs)-based methods, such as Mamba, efficiently capture long-range dependencies but may struggle with preserving fine anatomical details. To address these challenges, we propose a segment anything model based context guided mamba unet, SAMBA-UNet, a novel framework that integrates SAM’s context knowledge with Mamba’s hierarchical feature learning to enhance both global representation and local detail refinement. Specifically, we introduce the Hierarchical Context Enhancement Module (HCEM), an attention-based fusion module that deeply incorporates the multi-level features of SAM and Mamba to improve detail perception and segmentation precision. Additionally, we employ LoRA and Adapter-based fine-tuning to adapt SAM to medical imaging tasks, further enhancing its domain-specific segmentation capability and robustness. We conducted a comprehensive study on several medical segmentation datasets, including Synapse, ISIC17, and ISIC18, evaluating our method across multiple metrics. Experimental results demonstrate that SAMBA-UNet achieves state-of-the-art performance, effectively balancing generalization, segmentation accuracy, and computational efficiency. Our code will be released upon acceptance.

\end{abstract}


\end{frontmatter}
\section{Introduction}
\label{sec:intro}
Medical image segmentation plays a crucial role in various clinical applications, including CT and MRI analysis, enabling precise extraction of pathological regions such as tumors and lesions. This aids early disease detection and accurate diagnosis\cite{nc1,aids}. However, unlike natural images, medical images often exhibit complex anatomical structures and overlapping pathological tissues, making segmentation a challenging task. Traditional CNN-based models like U-Net\cite{unet} excel in capturing local texture and edge features, yet their limited receptive fields often fail to incorporate broader spatial relationships essential for accurate region delineation. To address this limitation, Transformer-based architectures introduced self-attention mechanisms capable of modeling global dependencies, enhancing their ability to integrate distant anatomical cues. However, their quadratic computational complexity significantly increases the burden when processing high-resolution or volumetric (3D) medical image data, often rendering them impractical for clinical deployment due to memory and inference constraints

In response, State Space Model (SSM) architectures, particularly the Mamba family, have emerged as an efficient and context-aware alternative. Inspired by structured state-space sequence models (e.g., S4), Mamba applies these principles to visual data, enabling long-range dependency modeling with linear complexity, while also being GPU-hardware-aware. In the medical domain, hybrid CNN–SSM networks like U‑Mamba combine convolutional layers for local texture and Mamba blocks for global context, demonstrating robust scalability across various imaging modalities. Pure SSM networks such as VM‑UNet and SegMamba further push this model paradigm by leveraging Visual State Space modules or multi-directional Mamba blocks to process dense volumetric data efficiently and coherently. A recent survey even reports that Mamba-based architectures outperform traditional transformer and CNN approaches in high-resolution image tasks while requiring fewer resources.

The Segment Anything Model (SAM), as a foundation model trained on large-scale natural image datasets, exhibits strong generalization and zero-shot segmentation capabilities. Its prompt-driven architecture allows for flexible object delineation and cross-domain adaptability. However, when applied directly to medical imaging, SAM often fails to accurately capture low-contrast anatomical boundaries or subtle pathological regions, due to domain-specific discrepancies in semantics and texture distribution. In contrast, Mamba-style State Space Models (SSMs) have shown great promise in medical image segmentation by efficiently modeling long-range dependencies with linear computational complexity. These models are particularly effective at capturing spatial continuity and structural context across volumetric data, yet they lack the rich semantic priors that foundation models like SAM provide—limiting their performance in cases of ambiguous or poorly contrasted anatomy. These observations suggest that SAM and Mamba offer highly complementary strengths: the former provides global semantic awareness, while the latter excels at structural consistency and context reasoning. A unified framework combining the two can potentially overcome the respective limitations of each, enabling segmentation models that are both semantically informed and anatomically precise. Despite this promising direction, prior research has mostly focused on either adapting SAM to medical domains or refining Mamba-based architectures independently. To date, no studies have systematically integrated SAM’s learned representations with Mamba’s context modeling to address the unique challenges of medical image segmentation—a gap this work aims to fill.

To bridge this gap, we propose SAMamba-UNet, a hybrid Mamba-based model that integrates the powerful visual representation learning of SAM with the long-range contextual modeling of Mamba. Our approach employs joint fine-tuning of SAM via LoRA\cite{LoRA} and Adapter\cite{Adapter} techniques to adapt it for medical image segmentation. Extensive experiments on Synapse\cite{synapsesite}, ISIC17\cite{ISIC17site}, and ISIC18\cite{ISIC18site} datasets demonstrate that SAMamba-UNet achieves competitive performance across various medical segmentation tasks, including organ and skin lesion segmentation. In summary, our primary contributions are as follows:
\begin{itemize}
\item[$1)$]We present SAMBA-UNet, a novel framework that integrates SAM’s contextual knowledge with Mamba’s hierarchical feature learning. By leveraging SAM’s generalization ability and Mamba’s long-range modeling, our method enhances global context understanding and fine-grained anatomical segmentation, achieving superior performance in medical image segmentation.
\item[$2)$]We designed a Hierarchical Context Enhancement Module (HCEM), a novel attention-based fusion module by deep fusion of SAM and mamba features, Use SAM's powerful generalization ability to enhance the model's perception of details.
\item[$3)$] We concurrently collaborative fine-tune SAM with the incorporation of both LoRA and Adapter to learn medical-specific knowledge, further enhance generalisation capabilities.
\item[$4)$] We conducte extensive experiments on three datasets, and the evaluation results show that our method exhibits competitive performance.
\end{itemize}

 \section{Related works}
 \subsection{Deep Neural Architectures for Medical Segmentation}
 Over the past decade\cite{nc2,nc3}, CNN-based U‑Net families—including Res‑UNet\cite{resunet}, Dense‑UNet\cite{denseunet}, UNet++\cite{UNet++}, Attention‑UNet\cite{attentionunet} and nnUNet\cite{nnunet}—have dominated medical image segmentation due to their effective local feature extraction and multi-scale representation capabilities. However, since their receptive field expands slowly, they inherently struggle to capture long-range dependencies. 
 
 To compensate for this, Transformers, starting from ViT and evolving into TransUNet\cite{transunet}, MedT\cite{jose2021medical}, SwinUNet\cite{swinunet}, and SwinUNETR\cite{swinunetr}, were introduced to incorporate global self-attention and capture distant anatomical structures. Yet, their quadratic computational complexity poses a significant burden when applied to high-resolution or 3D medical data. 
 
 More recently, Mamba-style State Space Models (SSMs) have emerged as a promising third paradigm. SSMs\cite{ssms} offer strong long-range dependency modeling with linear complexity, making them well-suited for dense prediction tasks. Notably, U‑Mamba\cite{U-Mamba}, a hybrid CNN‑SSM architecture, combines convolutions with Mamba blocks to achieve both local and global modeling in a scalable encoder–decoder framework, outperforming traditional CNNs and Transformers on multiple biomedical tasks like CT/MR organ, endoscopy instrument, and microscopy cell segmentation. Building on this, VM‑UNet\cite{Ruan2024Vm-unet} introduces a pure SSM-based U‑Net design by using Visual SSM blocks\cite{vmamba} and an asymmetric encoder–decoder architecture. Thus, the landscape of segmentation architectures has clearly evolved from local CNNs to global Transformers, and now to scalable, long-range SSMs like Mamba.

Despite the progress, most existing works treat these paradigms in isolation—either improving local context modeling via CNN variants or exploring global reasoning through Transformer attention. SSM-based designs offer a middle ground with favorable scalability, yet often lack high-level semantic guidance, especially in semantically sparse medical images. Furthermore, few works have investigated the fusion of state space modeling with external semantic priors, such as those from large-scale foundation models. This leaves open a promising direction for hybrid architectures that combine efficiency, spatial coherence, and semantic richness.
 
 \subsection{Foundation Models in Medical Segmentation}
In recent years, large-scale foundation models have begun to transform computer vision, with the Segment Anything Model (SAM)\cite{sam} emerging as a seminal example of prompt-based segmentation across diverse natural image datasets. However, when directly applied to medical imaging, SAM’s performance diminishes due to semantic and visual domain shifts that lead to inaccurate delineation of pathological regions. To bridge this gap, researchers have proposed several SAM-derived variants optimized for medical scenarios. Notably, MedSAM\cite{Medical sam adapter}, trained on over 1.5 million medical image–mask pairs spanning 10 modalities and more than 30 cancer types, demonstrated enhanced generalization across both internal and external datasets.However, these adaptations largely rely on prompt engineering or data re-training, and often do not address the need for structural consistency or efficient context modeling. While MedSAM introduces extensive medical priors via supervised pre-training, it remains prompt-dependent during inference and lacks explicit mechanisms for spatial alignment across slices or volumetric inputs.

 On the other hand, BioSAM‑2\cite{biosam}, based on SAM‑2\cite{sam2} and enhanced with memory-augmented mechanisms, addresses both single-frame 2D and multi-frame 3D biomedical segmentation tasks. Other techniques—including modalities augmentation in MA‑SAM\cite{msa}, which injects 3D adapter modules into SAM’s transformer blocks, and lightweight variants like MCP‑MedSAM\cite{mcpsam}, introducing modality and content prompts to reduce GPU requirements—have further improved applicability across imaging modalities. Many of these approaches focus on input-side adaptations—either via modality prompts, adapter injection, or multi-view learning—yet they seldom reconsider the downstream decoder design. In particular, the fusion mechanisms and structural modeling strategies are often left unchanged, limiting the network’s ability to align semantic features with anatomical details. Additionally, few methods incorporate mechanisms for capturing long-range dependencies across spatially distant regions within a single image, which is critical for segmenting dispersed or context-sensitive structures.

Despite these adaptations enhancing prompt automation, volumetric perception, and modality-specific tuning, most remain focused on input form or prompt engineering and have not integrated architectures designed for efficient long-range context modeling, such as Mamba. In contrast, our approach deliberately fills this gap. We first apply both LoRA\cite{lora} and Adapter\cite{msa} modules to fine-tune SAM\cite{sam}, instilling it with medical domain knowledge, and then fuse its powerful visual representations with a Mamba-based encoder–decoder backbone.
 \section{Methodology}
\subsection{Overview}
\begin{figure*}[t]
\centerline{\includegraphics[width=\textwidth]{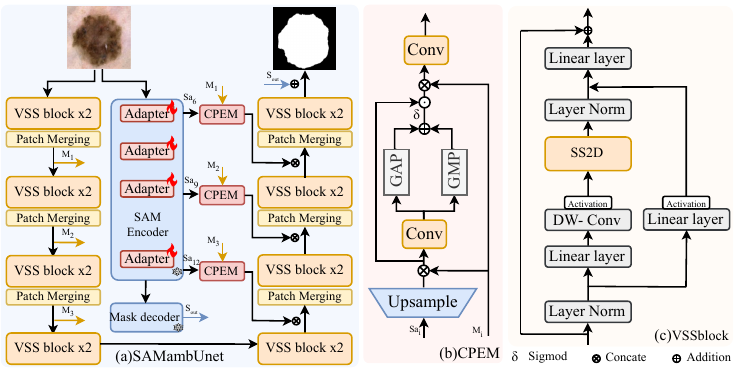}}
\caption{The overall structure of our proposed framework. }  
\label{fig1}
\end{figure*}

Existing vision Mamba models excel in long-range dependency modeling for general images but struggle with global contextual perception in medical imaging. To address this, we propose Segment Anything Model-Based Context Guided Mamba UNet (SAMBA-UNet), which leverages the Segment Anything Model (SAM)-a universal segmentation model with strong generalization and deep global context extraction via its encoder. First, we introduceCollaborative SAM Encoder Fine-tuning (CoSE), utilizing LoRA and Adapter to adapt the SAM encoder to medical imaging. Second, we integrate multi-level SAM encoder features as prior knowledge to enhance Mamba features during decoding. Furthermore, to fully exploit SAM’s rich contextual information and effectively guide Mamba feature learning, we propose a Hierarchical Context Enhancement Module (HCEM), as shown in Figure \ref{fig1}.
\subsection{SAMBA-UNet}
As illustrated in Figure \ref{fig1}-(a), we first apply patch embedding to the medical image, then simultaneously encode features using both the Mamba and SAM encoders. The Mamba encoder is constructed by deeply stacking VSS blocks, as illustrated in Figure \ref{fig1}-(c) while the SAM encoder undergoes fine-tuning with Collaborative LoRA and Adapter (CoSE) during training. Furthermore, we extract hierarchical features \( S_i \) (\( i = 6,9,12 \)) from the SAM encoder and skip connection features from the VSS blocks. These features are fused using our proposed Hierarchical Context Enhancement Module (HCEM), which performs multi-scale transformation and deep guidance fusion. This process leverages the rich contextual information in SAM features as prior knowledge to guide the subsequent Mamba decoding. Finally, a category-specific mapping is applied to generate the final segmentation result. Furthermore, since SAM utilizes a lightweight decoder to process the final-layer features, we can selectively incorporate SAM’s decoded features into the final prediction process based on performance evaluation.

\subsection{Collaborative Fine-Tuning of the SAM Encoder (CoSE)}
To adapt the Segment Anything Model (SAM) encoder to domain-specific medical imaging tasks while preserving its pre-trained knowledge, we propose a collaborative fine-tuning of the SAM Encoder (CoSE) that combines Low-Rank Adaptation (LoRA) and Adapter modules. This approach ensures efficient adaptation while preserving the pre-trained knowledge of SAM’s transformer-based encoder. Based on MedSAM \cite{Medical sam adapter}, we incorporate LoRA and Adapter into the Multi-Head Self-Attention (MHSA) and Feed-Forward Network (FFN) of each ViT block, respectively. First, we fine-tune the Query \( Q \) and Value \( V \) matrices of MHSA using LoRA. Specifically, as shown in the following equation:
\begin{equation}\label{eq:1}
Q^{'}, V^{'} = (Q, V) + W_{up}(W_{down}(Q, V))
\end{equation}
where \(W_{down}\) and \(W_{up}\) represent the linear down-projection and up-projection matrices in the LoRA mechanism, respectively. These transformations adjust the attention computation in the Multi-Head Self-Attention (MHSA) module to fine-tune the feature extraction for SKBS. Furthermore, we integrate two Adapters into each ViT block. The Adapter modules are formulated as follows:
\begin{equation}\label{eq:2}
X_{out} =  W_{up}(GELU(W_{down}(X_{in})))
\end{equation}
where \(W_{down}\) and \(W_{up}\) denote the down-projection and up-projection operations in the Adapter, respectively. The first Adapter is inserted after MHSA and before the residual connection, while the second is positioned within the residual path of the MLP layer. This integration effectively adapts the pre-trained ViT block to medical image segmentation tasks.

\subsection{Hierarchical Context Enhancement Module (HCEM)}  
To effectively integrate the global contextual knowledge from the Segment Anything Model (SAM) and the long-range dependency modeling ability of Mamba, we propose the Hierarchical Context Enhancement Module (HCEM). This module enables deep fusion of hierarchical SAM features \( S_{a_i} \) (\( i=6,9,12 \)) and Mamba skip connection features \( M_i \) (\( i=1,2,3 \)), ensuring robust feature alignment and enhancement. Given the difference in spatial resolutions between SAM hierarchical features and Mamba skip connection features, we first perform a multi-scale transformation on \( S_{a_i} \) to unify its scale with \( M_i \). This ensures that features from both sources are aligned spatially before fusion. The transformed SAM features and Mamba skip connection features are then concatenated along the channel dimension:

\begin{equation}
F_{\text{cat}} = \text{Concat}( \text{MS}(S_{a_i}), M_i )
\end{equation}

where \( \text{MS}(\cdot) \) denotes the multi-scale transformation applied to \( S_{a_i} \), and \( F_{\text{cat}} \) represents the concatenated feature map. A \( 1 \times 1 \) convolution is then applied to adjust the channel dimensions, facilitating smooth integration of both feature representations. To further refine the fused features, we introduce an attention mechanism based on both average pooling and max pooling operations. The pooled features are passed through an activation function to generate attention weights:

\begin{equation}
F_{\text{att}} = \sigma(\text{AvgPool}(F_{\text{cat}}) + \text{MaxPool}(F_{\text{cat}}))
\end{equation}

where \( \sigma(\cdot) \) is the activation function, ensuring non-linearity in the attention generation process. The attention-enhanced features are then applied to the original fused features via residual connections, ensuring effective feature adaptation. Finally, the refined fusion features are concatenated again with the original Mamba skip connection features \( M_i \), followed by a \( 1 \times 1 \) convolution to unify the output channels, ensuring compatibility with the subsequent decoding process. This hierarchical fusion strategy leverages SAM’s global contextual information to effectively guide the Mamba feature representation, enhancing spatial perception and structural consistency in the segmentation task.

\subsection{End-to-End Training}
We employ Binary Cross-Entropy and Dice loss (BceDice loss) for binary segmentation, and Cross-Entropy and Dice loss (CeDice loss) for multi-class segmentation, following VMUNet \cite{Ruan2024Vm-unet}. The loss functions are defined as:

\begin{equation}\label{eq:5}
L_{BceDice} = \lambda_{1} L_{Bce} + \lambda_{2} L_{Dice}
\end{equation}

\begin{equation}\label{eq:6}
L_{CeDice} = \lambda_{3} L_{Ce} + \lambda_{4} L_{Dice}
\end{equation}

The values of $\lambda$ are set identical to those in VMUNet \cite{Ruan2024Vm-unet} to ensure fair comparisons.

\begin{table*}[htbp]
    \centering
        \renewcommand{\arraystretch}{0.9}
\caption{Comparative experimental results on the ISIC17 and ISIC18 dataset. (Bold indicates the best.)}
        \resizebox{0.8\linewidth}{!}{
\begin{tabular}{c|c|ccccc}
\hline
Dataset & Model & mIoU(\%)$\uparrow$ & DSC(\%)$\uparrow$ & ACC(\%)$\uparrow$ & Spe(\%)$\uparrow$ & Sen(\%)$\uparrow$ \\\hline
        & \multicolumn{1}{c|}{MedSAM\cite{Medical sam adapter}} &  75.48&
    84.15&
    93.34&
    98.04&
    80.20
    \\
    &\multicolumn{1}{c|}{IMedSAM\cite{IMedSAM}}     & 75.40  & 85.98 & 95.52 & 98.23 & 82.01 \\
        & \multicolumn{1}{c|}{Unet\cite{unet} }&76.98 &
    86.99&
    95.65&
    97.43&
    86.82\\
        &\multicolumn{1}{c|}{UTNetV2\cite{UTNetV2}} & 77.35 &
    87.23&
    95.84&
    98.05&
    84.85\\
         &\multicolumn{1}{c|}{HC-Mamba\cite{HC-Mamba}} & 79.27 & 88.18 & 95.01 & 97.47 & 86.62\\
       
        &\multicolumn{1}{c|}{TransFuse\cite{zhang2021transfuse}} & 79.21 &
    88.40&
    96.17&
    97.98&
    \textbf{97.14}\\
ISIC17  &\multicolumn{1}{c|}{MALUNet\cite{MALUNet}} & 78.78 &
    88.13&
    96.18&
    \textbf{98.47}&
    84.78\\
        &\multicolumn{1}{c|}{VM-UNet\cite{Ruan2024Vm-unet}} &     80.23 &
    89.03&
    96.29&
    97.58&
    89.90\\

        &\multicolumn{1}{c|}{Our} & \textbf{81.07}&
    \textbf{89.54}&
    \textbf{96.52}&
    98.08&
    88.81
    \\
        \hline
\hline
       & Unet \cite{unet}&
    77.86 &
    87.55&
    94.05&
    96.69&
    85.86\\
       &     UNet++\cite{UNet++} &
    78.31 &
    87.83&
    \textbf{94.02}&
    95.75&
    88.65\\
       &Att-UNet\cite{attentionunet} &
    78.43 &
    87.91&
    94.13&
    96.23&
    87.60\\
       & UTNetV2 \cite{UTNetV2}&
    78.97 &
    88.25&
    94.32&
    96.48&
    87.60\\
ISIC18 &  SANet\cite{Wei2021Shallow} &
    79.52 &
    88.59&
    94.39&
    95.97&
    89.46\\
    & MedSAM\cite{Medical sam adapter} &
    80.50&
    89.19&
    95.27&
    94.99&
    \textbf{96.37}    \\
       &  TransFuse \cite{zhang2021transfuse}&
    80.63 &
    89.27&
    94.66&
    95.74&
    91.28\\
       & VM-UNet\cite{Ruan2024Vm-unet} &
    81.35 &
    89.71&
    94.91&
    96.13&
    91.12\\

       &Ours &\textbf{81.69}
     &\textbf{89.92}
    &95.10
    &\textbf{96.85}
    &89.69
    \\ \hline  
\end{tabular}
  
    }
    \label{tab:1}
\end{table*}

\section{Experiment}
\subsection{Dataset and metrics}
\subsubsection{Dataset}
 To validate the effectiveness of our method, we conducted comprehensive experiments on three different datasets. Specifically, we evaluated the performance of SAMamba-UNet on medical image segmentation tasks in the Synapse\cite{synapsesite}, ISIC17\cite{ISIC17site}, and ISIC18\cite{ISIC18site} datasets.

\textbf{ISIC2017}: The ISIC2017 dataset contains 2,150 dermoscopic images with segmentation mask labels for three diseases: melanoma, seborrheic keratosis, and benign nevi. Additionally, there are 600 test images. The image sizes range from 767×576 to 6621×4441.

 \textbf{ISIC2018}: The ISIC2018 dataset contains a different number of lesion images for classification and segmentation tasks, with 2,694 dermoscopic images having segmentation mask labels for segmentation.

 \textbf{Synapse}: The Synapse Multi-Organ Segmentation dataset is a publicly available multi-organ segmentation dataset, containing 30 abdominal CT cases with 3,779 longitudinal clinical abdominal CT images. These images have uniform resolution and provide accurate trachea contour annotations.
 
 \textbf{Metrics} To compare our method with other state-of-the-art approaches, we maintained consistency in evaluation metrics. For the ISIC2017 and ISIC2018 datasets, we used mean Intersection over Union (mIoU), Dice Similarity Coefficient (DSC), Accuracy (Acc), Sensitivity (Sen), and Specificity (Spe) as the evaluation metrics. For the Synapse dataset, we used Dice Similarity Coefficient (DSC) and the 95\% Hausdorff Distance (HD95) as evaluation metrics.

\subsection{Implementation Details}  
All images were resized to $256 \times 256$ to adapt to the VMamba-S model. The batch size was set to 32, and training was conducted using the AdamW optimizer with an initial learning rate of $1e^{-3}$. A CosineAnnealingLR scheduler was used, with a maximum iteration of 50 and a minimum learning rate of $1e^{-4}$ for Synapse dataset. Training was performed for 300 epochs using VMamba-S\cite{vmamba}\footnote{Pre-trained weights: vmamba\_small\_e238\_ema.pth} as the pre-trained model. All experiments were conducted on an NVIDIA RTX A6000 GPU.

\begin{table*}[]
    \centering
    \caption{Comparative experimental results on the Synapse dataset. DSC of every single class is also reported. (Bold indicates the best.)}
        \resizebox{1\linewidth}{!}{
    \begin{tabular}{c|cc|cccccccc}
         \hline
         \textbf{Model}&
         \textbf{DSC$\uparrow$}&
         \textbf{HD95}$\downarrow$ &
         \textbf{Aorta} &
         \textbf{Gallbladder} &
         \textbf{Kidney(L)} &
         \textbf{Kidney(R)} &
          \textbf{Liver}&
          \textbf{Pancreas}&
          \textbf{Spleen}&
          \textbf{Stomach} \\ \hline 
      
          MedSAM\cite{Medical sam adapter}&
          73.99&
          23.68&
          67.15&
          60.99&
          67.68&
          66.79&
          88.92&
          70.72&
          87.20&
          82.48\\

          U-Net\cite{unet}&
          76.85&
          39.70&
          89.07&
          \textbf{69.72}&
          77.77&
          68.60&
          93.43&
          \textbf{53.98}&
          86.67&
          75.58\\
          Att-UNet\cite{attentionunet}&
          77.77&
          36.02&
          89.55&
          68.88&
          77.98&
          71.11&
          93.57&
          58.04&
          87.30&
          75.75\\
     
          TransUNet\cite{transunet}&
          77.48&
          31.69&
          87.23&
          63.13&
          81.87&
          77.02&
          94.08&
          55.86&
          \textbf{85.08}&
          75.62\\
       
          TransDeepLab\cite{Azad2022Transdeeplab}&
          80.16&
          21.25&
          86.04&
          69.16&
          84.08&
          79.88&
          93.53&
          61.19&
          89.00&
          78.40\\
          UCTransNet\cite{uctransunet}&
          78.23&
          26.75&
          -&
          -&
          -&
          -&
          -&
          -&
          -&
          -\\

          MEW-UNet\cite{Ruan2022MEW-UNet}&
          78.92&
          \textbf{16.44}&
          86.68&
          65.32&
          82.87&
          80.02&
          93.63&
          58.36&
          90.19&
          74.26\\
          SAM3D\cite{sam3d}&
          79.56&
          17.87&
           85.64&
          86.31&
         84.29&
          49.81 &
        76.11&
        69.32&
          89.57&
          95.42\\

          MedMamba\cite{Yubiao2024Medmamba}&
          80.67&
          28.15&
          \textbf{88.23}&
          \textbf{69.53}&
          82.22&
          79.49&
          \textbf{96.89}&
          58.57&
          92.29&
          79.12\\          
          VM-UNet\cite{Ruan2024Vm-unet}&
          81.08&
          19.21&
          86.40&
          69.41&
          86.16&
          82.76&
          94.17&
          58.80&
          89.51&
          \textbf{81.40}\\
          Ours&
          \textbf{81.94}&
          17.34&
          87.12&
         68.68&
          \textbf{86.80}&
          \textbf{83.91}&
          94.36&
          64.59&
          89.77&
          80.26
          \\\hline

    \end{tabular}
            }
    \label{tab:2}
\end{table*}

\subsection{Comparison with State-of-the-art Methods}
To further validate the effectiveness and generalization ability of our SAMBA-UNet framework, we comprehensively compare it against a wide range of state-of-the-art (SOTA) medical image segmentation methods, including CNN-based, Transformer-based, SSM-based, and SAM-derived foundation models. The evaluation is conducted on three benchmark datasets: ISIC17, ISIC18, and Synapse.
\newline\textbf{Comparison experiments on ISIC17 and ISIC18.}The experimental results on the ISIC17 and ISIC18 datasets, as shown in Table \ref{tab:1}, demonstrate the superior performance of our proposed method compared to existing approaches. On ISIC17, our method achieves the highest mIoU 81.07\% and DSC 89.54\%, outperforming the previous best-performing VM-UNet by 0.84\% and 0.51\%, respectively. Similarly, on ISIC18, our model achieves the best mIoU 81.69\% and DSC 89.92\%, surpassing VM-UNet by 0.34\% and 0.21\%. We believe that traditional CNN‑based architectures such as U‑Net, UNet++, and Attention‑U‑Net capture local textures effectively but fall short in modeling long‑range contextual relationships. This often leads to incomplete or blurred segmentation of diffuse lesion borders and results in false positives—issues which hybrid Transformer‑CNN models like TransUNet only partially address. Although TransUNet outperforms pure CNNs (e.g., reaching around 89.8\% Dice on ISIC‑17), it still struggles with preserving sharp lesion boundaries and involves heavier computation. Meanwhile, prompt‑dependent foundational models such as MedSAM, despite their strong pretrained capabilities, lack robustness in fully automatic scenarios—often missing small irregular lesions due to prompt sensitivity and inability to adapt boundary precision across samples. By contrast, SAMBA‑UNet is designed to overcome these limitations: its Mamba-SSM modules capture global relational patterns across lesion areas, while its HCEM-augmented decoder refines local boundaries to ensure crisp, accurate masks. Crucially, SAMBA‑UNet operates in a prompt‑free manner, enabling consistent and automated segmentation across the entire dataset. As a result, it consistently outperforms these methods on ISIC17/18 in both quantitative metrics and visual fidelity. These results indicate that SAMBA-UNet achieves a favorable balance between generalization and precision. Its prompt-free structure avoids reliance on handcrafted inputs, while the hierarchical integration of contextual features contributes to accurate boundary delineation. We also observe consistent performance across different lesion types and sizes, highlighting the method’s robustness under varying clinical conditions.


\par
\begin{figure*}[t]
\centering
\includegraphics[width=1\textwidth]{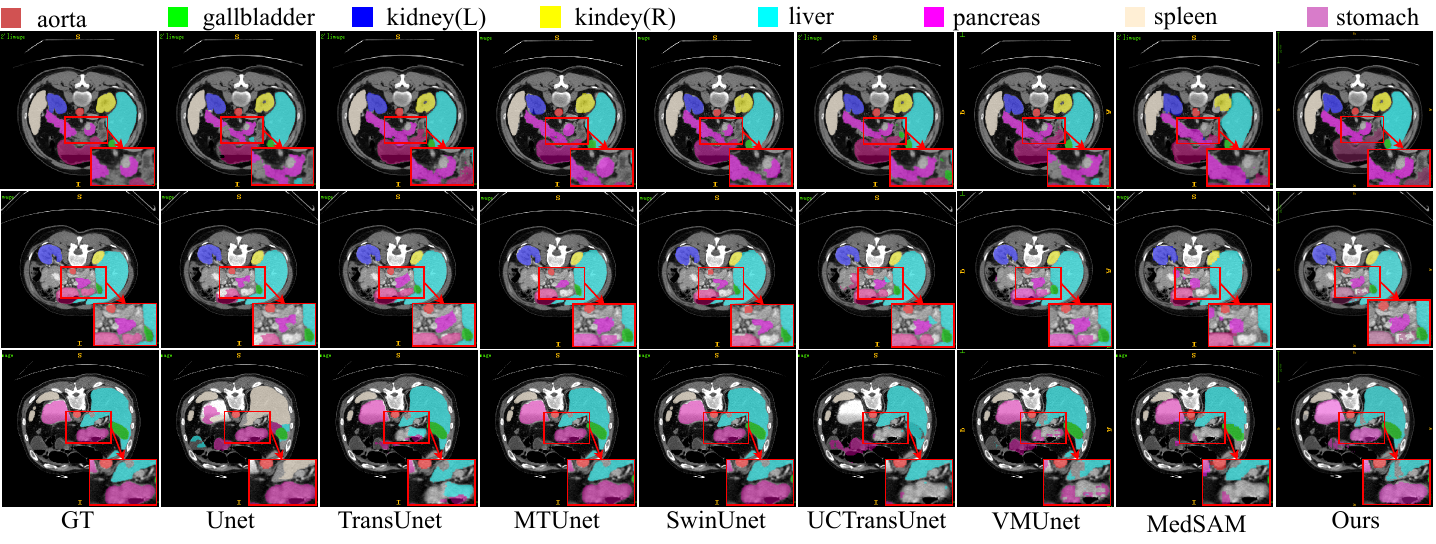}
\caption{ Visualization results on the Synapse datasets.  } 
\label{fig2}
\end{figure*}

\noindent\textbf{Comparison experiments on Synapse.}
As shown in Table \ref{tab:2}, our method achieves the highest DSC 81.94\%, surpassing the previous best-performing VM-UNet by 0.86\%. Additionally, our method attains the lowest HD95 17.34, demonstrating improved segmentation boundary precision. In terms of organ-specific segmentation, our model outperforms others on the left kidney 86.80\% and right kidney 83.91\%, while maintaining competitive performance across other anatomical structures. The visualization results are shown in Figure \ref{fig2}. Similarly, on the Synapse multi‑organ CT dataset, pure CNN models fail to fully leverage spatial relationships among adjacent organs, often missegmenting overlapping structures. Pure Transformer approaches such as Swin‑UNet add global reasoning but produce coarse organ boundaries and incur high computational costs. For instance, hybrid Mamba‑based models like VM‑UNet achieve only ~82\% DSC and ~19mm HD95, while recent ACM‑UNet reaches ~85.12\% DSC and ~13.9mm HD95. However, these models still demand substantial compute resources. Additionally, SAM‑based models do not generalize well in 3D: their prompt‑centric designs lead to slice‐inconsistent predictions and fail to maintain accurate delineations across volumetric data. SAMBA‑UNet addresses all these shortcomings simultaneously: it achieves efficient global context modeling via Mamba‑SSM, while HCEM-driven decoding ensures boundary consistency across slices. Prompt‑free segmentation removes manual dependency and improves volumetric cohesion, and LoRA plus Adapter fine‑tuning preserves pretrained knowledge with minimal parameters. Consequently, on Synapse, SAMBA‑UNet delivers higher DSC and lower HD95 than competing paradigms, producing smoother, anatomically accurate segmentations even for challenging organs like pancreas and gallbladder. In sum, across both ISIC and Synapse datasets, competing approaches consistently suffer from at least one significant limitation—CNNs lack global context, Transformers yield coarse boundaries or demand heavy compute, and SAM-based methods falter with prompt reliance or domain inconsistency. By contrast, SAMBA-UNet integrates the global semantic awareness of foundation models with the efficient contextual modeling of Mamba, enabling accurate, prompt-free segmentation across diverse anatomical structures. Its ability to maintain high accuracy, edge fidelity, and spatial coherence—without reliance on manual prompts or post-processing—makes it particularly well-suited for deployment in automated clinical workflows.
﻿
\begin{table*}[t]
  \centering
        \caption{Ablation study of our method on the Synapse datasets.}
      \resizebox{0.6\linewidth}{!}{
    \begin{tabular}{c|ccc|cc}
    \toprule
       &CoSE (Lora)  &CoSE (Adapter) & HCEM  & DSC$\downarrow$  & HD95$\uparrow$ \\
    \midrule
    (a)   &    &       &       & 79.05 & 29.89 \\
    (b)   &  \checkmark        & &          & 80.33 & 23.69 \\
    (c)   &       &\checkmark    &       & 80.58 & 24.92\\
    (d)   & \checkmark     & \checkmark     &        & 81.06 & 23.18  \\
    (e)   & \checkmark     & \checkmark     & \checkmark         & 81.94 & 17.34 \\
    \bottomrule
    \end{tabular}%
      }
  \label{tab:3}%
\end{table*}%

\subsection{Ablation Studies} 

As shown in Table \ref{tab:3}, our ablation study evaluates the impact of different components in our framework. Starting from the baseline model (a), incorporating LoRA-based fine-tuning (b) improves DSC by 1.28\% and reduces HD95 by 6.2\%, while using only Adapter-based fine-tuning (c) further enhances DSC to 80.58\% and reduces HD95 to 24.92. Combining both LoRA and Adapter in CoSE (d) achieves a DSC of 81.06\% with an HD95 of 23.18\%. This indicates that the two tuning methods offer complementary benefits: LoRA enhances the attention weights within the transformer encoder, while Adapter introduces additional task-specific capacity without altering the original parameters. Their joint use leads to more robust feature adaptation for the medical domain.

Additionally, it can be seen in integrating HCEM experiment (e) achieves a DSC of 81.94\% with an HD95 of 17.34\%, demonstrating its effectiveness in enhancing segmentation. The performance gain from HCEM suggests that high-level semantic priors extracted from the SAM encoder can effectively guide the Mamba decoder via deep hierarchical fusion. By aligning semantic and structural features across multiple levels, HCEM facilitates more precise boundary localization and improves spatial consistency. 

Overall, the ablation study confirms that each component in our framework contributes meaningfully to final performance, and the full configuration delivers the best results in both accuracy and boundary precision.


\section{Conclusion}
We propose SAMBA-UNet, a hybrid segmentation framework that combines SAM’s contextual knowledge with Mamba’s hierarchical feature modeling for medical image segmentation. By introducing the Hierarchical Context Enhancement Module (HCEM) and applying LoRA and Adapter-based collaborative fine-tuning, our model effectively leverages global semantic priors while maintaining domain-specific adaptability and structural precision.

Extensive experiments demonstrate that SAMBA-UNet achieves superior accuracy, boundary fidelity, and robustness across multiple datasets, all without relying on prompt-based inputs. Its prompt-free and modular design makes it suitable for integration into real-world clinical segmentation pipelines. In future work, we aim to explore mamba-based lightweight tuning strategies to further optimize computational efficiency and extend our framework to broader imaging modalities.


\end{document}